\documentclass[nohyperref]{article}

\usepackage{microtype}
\usepackage{graphicx}
\usepackage{subfigure}
\usepackage{booktabs} 

\usepackage{hyperref}

\usepackage[accepted]{icml2022}

\usepackage{amsmath}
\usepackage{amssymb}
\usepackage{mathtools}
\usepackage{amsthm}
\usepackage{amssymb}
\usepackage{latexsym}
\usepackage{graphicx}
\usepackage{amsfonts}
\usepackage{amsmath}
\usepackage{booktabs}
\usepackage{tabularx}
\usepackage{algorithm}
\usepackage{enumitem}
\usepackage{xcolor}
\usepackage{xspace}

\usepackage[capitalize,noabbrev]{cleveref}

\theoremstyle{plain}
\newtheorem{theorem}{Theorem}[section]

\theoremstyle{definition}
\newtheorem{definition}[theorem]{Definition}

\theoremstyle{remark}

\newcommand{\vfamily}{$\mathcal{V}$\xspace}
\newcommand{\vi}{$\mathcal{V}$-information\xspace}
\newcommand{\pvi}{\textsc{pvi}\xspace}
\newcommand{\pmi}{\textsc{pmi}\xspace}

\newcommand{\vinfo}{$I_\mathcal{V}(X \to Y)$\xspace}

\newcommand{\draftonly}[1]{#1}
\newcommand{\draftcomment}[1]{\draftonly{#1}}

\newcommand{\kawin}[1]{\draftcomment{{\color{red}[#1]$_{ke}$}}}

\newcommand{\longtermtodos}[1]{\draftcomment{{\color{orange}[#1]$_{CamReady-TODO}$}}}

\renewcommand{\longtermtodos}[1]{}    

\usepackage[textsize=tiny]{todonotes}

\icmltitlerunning{Understanding Dataset Difficulty with $\mathcal{V}$-Usable Information}

\begin{document}

\twocolumn[
\icmltitle{Understanding Dataset Difficulty with $\mathcal{V}$-Usable Information}



\icmlsetsymbol{equal}{}

\begin{icmlauthorlist}
\icmlauthor{Kawin Ethayarajh}{stanford}
\icmlauthor{Yejin Choi}{allenai,uw}
\icmlauthor{Swabha Swayamdipta}{allenai}
\end{icmlauthorlist}

\icmlaffiliation{stanford}{Stanford University}
\icmlaffiliation{allenai}{Allen Institute for Artificial Intelligence}
\icmlaffiliation{uw}{Paul G. Allen School of Computer Science, University of Washington}

\icmlcorrespondingauthor{Kawin Ethayarajh}{kawin@stanford.edu}

\icmlkeywords{Machine Learning, ICML}

\vskip 0.3in
]



\printAffiliationsAndNotice{Work done during an internship at AI2.} 

\begin{abstract}
Estimating the difficulty of a dataset typically involves comparing state-of-the-art models to humans; the bigger the performance gap, the harder the dataset is said to be.
However, this comparison provides little understanding of how difficult each instance in a given distribution is, or what attributes make the dataset difficult for a given model.
To address these questions, we frame dataset difficulty---w.r.t.\ a model $\mathcal{V}$---as the lack of $\mathcal{V}$-\emph{usable information} \citep{xu2019theory}, where a lower value indicates a more difficult dataset for $\mathcal{V}$.
We further introduce \emph{pointwise $\mathcal{V}$-information} (\pvi) for measuring the difficulty of individual instances w.r.t.\ a given distribution.
While standard evaluation metrics typically only compare different models for the same dataset, $\mathcal{V}$-\emph{usable information} and \pvi also permit the converse: for a given model $\mathcal{V}$, we can compare different datasets, as well as different instances/slices of the same dataset.
Furthermore, our framework allows for the interpretability of different input attributes via transformations of the input, which we use to discover annotation artefacts in widely-used NLP benchmarks. 
\end{abstract}
\section{Introduction}
\label{sec:intro}

Datasets are designed to act as proxies for real-world tasks, yet most bear limited semblance to the tasks they purport to reflect \citep{torralba2011unbiased,recht2019imagenet}.
Understanding dataset difficulty is therefore imperative to understanding progress in AI.
In practice, however, estimating dataset difficulty is often limited to an informal comparison of state-of-the-art model performance to that of humans; the bigger the performance gap, the harder the dataset is said to be \citep{ethayarajh2020utility,ma2021dynaboard}.
However, such performance metrics offer little understanding of the differential difficulty of individual instances, or of which attributes in the input a given model finds useful.

\begin{figure}
    \centering
    \includegraphics[width=\columnwidth]{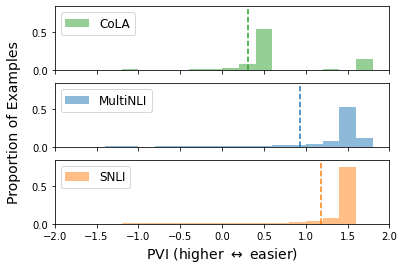}
    \caption{
    The Stanford NLI dataset contains more BERT-usable information than the MultiNLI and CoLA datasets, making it easier for BERT-base.
    Above, the distribution of instance difficulty (\pvi) in the held-out sets for each; dotted lines denote the average \pvi. 
    }
    \label{fig:teaser}
\end{figure}

To understand why a dataset is difficult, we extend recent work in information theory \cite{xu2019theory}.
To illustrate, consider a model family $\mathcal{V}$ that can learn to map a sentence $X$ with its sentiment $Y$.
Even if $X$ were to be encrypted, the information $X$ contains about $Y$ would not be removed; in other words, the Shannon mutual information 
would be unchanged \cite{shannon1948mathematical}.
However, encryption makes predicting the sentiment a lot more difficult for $\mathcal{V}$.
But why? 
Intuitively, the task is easier when $X$ is unencrypted because the information it contains is \emph{usable} by $\mathcal{V}$; when $X$ is encrypted, the information still exists but becomes unusable.
This quantity---$\mathcal{V}$-\emph{usable information}---reflects the ease with which $\mathcal{V}$ can predict $Y$ given $X$.
\citet{xu2019theory} show that it can be measured using the \emph{predictive $\mathcal{V}$-information} framework, which generalizes Shannon information to consider computational constraints.

Our work extends the above framework by framing dataset difficulty as the lack of $\mathcal{V}$-\emph{usable information}.\footnote{We use the terms ``{$\mathcal{V}$-usable information}'' and ``{$\mathcal{V}$-information}'' from \citet{xu2019theory}, interchangeably.}
The higher the $\mathcal{V}$-usable information, the easier the dataset is for $\mathcal{V}$.
Not only does this framework allow comparisons of models w.r.t.\ the same dataset, but also of different datasets w.r.t.\ the same model. 
Figure \ref{fig:teaser} illustrates that different datasets provide different amounts of usable information for the same model, even when the task is identical (i.e., natural language inference in the SNLI \cite{bowman2015large} and MultiNLI \cite{williams2018broad} datasets).

Building on the aggregate estimate of dataset difficulty, we introduce a measure called \emph{pointwise $\mathcal{V}$-information} (\pvi) for estimating the difficulty of each instance w.r.t.\ a given distribution (\S\ref{sec:pvi}).
\pvi estimates allow us to compare not only individual instances, but also the difficulty of slices of data w.r.t\ $\mathcal{V}$. 
On datasets containing more usable information (e.g., SNLI), \pvi estimates are highly correlated (Pearson $r \geq$ 0.75) across different models, seeds, and training time, and with human judgments of difficulty.

Comparisons of $\mathcal{V}$-usable information before and after isolating an input attribute shed light on \emph{why} the dataset is easy or difficult for $\mathcal{V}$ (\S\ref{sec:understanding}), which has significant implications for interpretability in AI\footnote{Our code and data are available \href{https://github.com/kawine/dataset_difficulty}{here}.} \cite{miller2019explanation}.
Specifically, we use $\mathcal{V}$-usable information to identify some limitations in benchmarks that are widely used in NLP to test for a model's understanding of different language phenomena:

\begin{itemize}
    \setlength\itemsep{0em}
    \item Word ordering has a limited impact on the difficulty of a popular natural language entailment benchmark, SNLI \citep{bowman2015large}, even though entailment describes a causal relationship.
    
    \item Some of the most difficult instances in SNLI and a popular grammaticality detection benchmark, CoLA \citep{warstadt2018neural}, are mislabelled.
    
    \item In a popular dataset for hate speech detection \cite{hateoffensive}, just 50 (potentially) offensive words contain most of the BERT-usable information about the label; less subtle bias may be going undetected.
\end{itemize}

\section{$\mathcal{V}$-Usable Information}
\label{sec:v-info}

\subsection{Background}

Consider a model family $\mathcal{V}$, which can be trained to map text input $X$ to its label $Y$.
If we encrypted the text, or translated it into a language with a very complex grammar, it would be harder to predict $Y$ given $X$ using the \textit{same} \vfamily. 
How might we measure this increase in difficulty?
\citet{shannon1948mathematical}'s mutual information $I(X;Y)$ is not an option---it would not change after $X$ is encrypted, as it allows for unbounded computation, including any needed to decrypt the text. 

Intuitively, the task is easier when $X$ is \textit{unencrypted} because the information it contains is \emph{usable} by \vfamily; when $X$ is encrypted, this information still exists but becomes unusable.
This quantity, called \textbf{$\mathcal{V}$-usable information}, provides an estimate of the difficulty of a dataset w.r.t. $\mathcal{V}$.
It can be measured under a framework called \textbf{predictive $\mathcal{V}$-information}, which generalizes Shannon information to measure how much information can be extracted from $X$ about $Y$ when constrained to functions \vfamily, written as \vinfo \citep{xu2019theory}.
The greater \vinfo is, the easier the dataset is for $\mathcal{V}$.
If $\mathcal{V}$ is the set of all functions---i.e., under unbounded computation---$\mathcal{V}$-information reduces to Shannon information.

Processing the input with $\tau$ (e.g., by decrypting the text) can make prediction easier, allowing $I_\mathcal{V}(\tau(X) \to Y) \geq I_\mathcal{V}(X \to Y)$.
Although this violates the data processing inequality, it explains the usefulness of certain types of processing, such as representation learning.
Compared to $X$, the learned representations cannot have more Shannon information with $Y$, but they can have more usable information. 

\subsection{Definitions}
As defined in \citet{xu2019theory}:

\begin{definition}
Let $X, Y$ denote random variables with sample spaces $\mathcal{X}, \mathcal{Y}$ respectively. 
Let $\varnothing$ denote a null input that provides no information about $Y$. 
Given predictive family $\mathcal{V} \subseteq \Omega = \{ f: \mathcal{X} \cup \varnothing \to P(\mathcal{Y}) \}$, the \textbf{predictive $\mathcal{V}$-entropy} is 
\begin{equation}
    H_\mathcal{V}(Y) = \inf_{f \in \mathcal{V}} \mathbb{E} [- \log_2 f[\varnothing](Y) ]
    \label{v-entropy}
\end{equation}
and the \textbf{conditional $\mathcal{V}$-entropy} is 
\begin{equation}
    H_\mathcal{V}(Y|X) = \inf_{f \in \mathcal{V}} \mathbb{E} [- \log_2 f[X](Y) ]
    \label{cond-v-entropy}
\end{equation}
We use $\log_2$ to measure the entropies in bits of information, though one could also use $\log_e$ and measure them in nats instead.
\end{definition}

Put simply, $f[X]$ and $f[\varnothing]$ produce a probability distribution over the labels.
The goal is to find the $f \in \mathcal{V}$ that maximizes the log-likelihood of the label data with (Eq. \ref{cond-v-entropy}) and without the input (Eq. \ref{v-entropy}). 
$f[\varnothing]$ models the label entropy, so $\varnothing$ can be set to an empty string for most NLP tasks.
Although \emph{predictive family} has a technical definition\footnote{$\mathcal{V}$ is a subset of all possible mappings from $\mathcal{X}$ to $P(\mathcal{Y})$ that satisfies \emph{optional ignorance}: for any $P$ in the range of some $f \in \mathcal{V}$, there exists some $f' \in \mathcal{V}$ s.t.\ $f'[X] = f'[\emptyset] = P$. See \citet{xu2019theory} for why optional ignorance is necessary.}, most neural models, provided they are finetuned without any frozen parameters, easily meet this definition.
Further, as per \citet{xu2019theory}:

\begin{definition}
Let $X$ and $Y$ denote random variables with sample spaces $\mathcal{X}$ and $\mathcal{Y}$, respectively. 
Given a predictive family \vfamily, the \textbf{$\mathcal{V}$-information} is 
\begin{equation}
    I_\mathcal{V}(X \to Y) = H_\mathcal{V}(Y) - H_\mathcal{V}(Y|X)
    \label{eq:v-info}
\end{equation}
\end{definition}
Because we are estimating this quantity on a finite dataset, the estimate can differ from the true $\mathcal{V}$-information. 
\citet{xu2019theory} provide PAC bounds for this error, where less complex \vfamily and larger datasets yield tighter bounds. 
 \citet{xu2019theory} also list several useful properties of $\mathcal{V}$-information:
\begin{itemize}[noitemsep,topsep=0pt]
    \item \emph{Non-Negativity:} $I_\mathcal{V}(X \to Y) \geq 0$
    \item \emph{Independence:} If $X$ is independent of $Y$, $I_\mathcal{V}(X \to Y) = 0$.
    \item \emph{Montonicity:} If $\mathcal{U} \subseteq \mathcal{V}$, then $H_\mathcal{U}(Y) \geq H_\mathcal{V}(Y)$ and $H_\mathcal{U}(Y|X) \geq H_\mathcal{V}(Y|X)$.
\end{itemize}

Training with the cross-entropy loss finds the $f \in \mathcal{V}$ that maximizes the log-likelihood of $Y$ given $X$ \citep{xu2019theory}.
Thus, $H_\mathcal{V}(Y|X)$ can be easily computed by standard training or by finetuning a pre-trained model.\footnote{
Improving model calibration using more advanced methods \citep{kumar2019verified} is a possible direction of future work.}
We estimate $H_\mathcal{V}(Y|X)$ by calculating $\mathbb{E}[- \log f[X](Y)]$ on an identically distributed held-out set, where $Y$ is the gold label.
Since training with cross-entropy ultimately aims to find the infimum over the \emph{data distribution}, not just the training set, it is important not to overfit the model to the training instances; this is of added significance for estimating $H_\mathcal{V}(Y|X)$.
We estimate $H_\mathcal{V}(Y)$ by training or finetuning another model where $X$ is replaced by $\varnothing$, intended to fit the label distribution.
As such, estimating $\mathcal{V}$-information involves training or finetuning only two models.

\subsection{Assumptions} 

Implicit in estimating the $\mathcal{V}$-information is the assumption that the data used to find the optimal $f \in \mathcal{V}$ and the data used to estimate $H_\mathcal{V}(Y|X)$ are identically distributed.
This dependence on the data distribution makes $\mathcal{V}$-information well-suited for estimating and interpreting dataset difficulty.
However, it is still possible to estimate the difficulty of sub-populations or subsets of the data, though it would be imprecise to refer to this measure as \vi (see \S\ref{ssec:second_implications} for details).
We also assume that the difference between the empirical \vi (calculated using some finite dataset) and the true \vi (calculated over the distributions) is negligible, though this may not hold, for example, if the dataset is too small (see Appendix \ref{appendix:trainingdata}). 

\begin{figure}[t]
    \centering
    \includegraphics[width=\columnwidth]{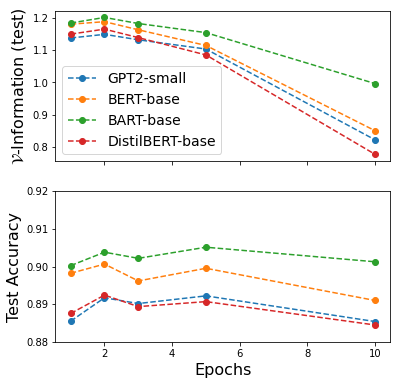}
    \vskip \bigskipamount
    \caption{Comparing the $\mathcal{V}$-usable information estimate to accuracy in SNLI.
    In the first three epochs, estimates on the test set are similar across all models (top), but due to over-fitting, the estimates diverge and decline.
    The test accuracy (bottom) for each model loosely tracks the $\mathcal{V}$-information estimate for that model, since extracting information makes prediction easier.
    }
    \label{fig:vinfo-vs-accuracy}
\end{figure}

\subsection{Implications} 
\label{sec:implications}

$\mathcal{V}$-usable information allows us to compare
\begin{itemize}[noitemsep,topsep=0pt]
    \item[i.] different models \vfamily by computing \vinfo for the same $X,Y$ (Fig.~\ref{fig:vinfo-vs-accuracy}), 
    
    \item[ii.] different datasets $\{(x,y)\}$ by computing \vinfo for the same \vfamily (Fig.~\ref{fig:teaser}), and
    
    \item[iii.] different input variables $X_i$ by computing $I_\mathcal{V}(X_i \to Y)$ for the same $\mathcal{V}$ and $Y$ (Fig.~\ref{fig:attributes}; \S\ref{sec:understanding}).
\end{itemize}
While common classification metrics, such as accuracy or $F_1$ score, are often used for the above comparisons, $\mathcal{V}$-usable information offers a theoretically rigorous framework, making it better suited for interpretability.
The $\mathcal{V}$-usable information is measured in bits / nats (depending on the log base), allowing for standardized comparisons across models and datasets.
Additionally, consider the case where $X$ and $Y$ are independent: here, model accuracy would be no greater than the majority class frequency, but this frequency varies across datasets.
$\mathcal{V}$-information avoids this problem by factoring in the label entropy $H_\mathcal{V}(Y)$; if $X, Y$ are independent, then the $\mathcal{V}$-information is \textit{provably} zero.

Say we wish to compare two predictive families, $\mathcal{V}$ and $\mathcal{U}$, such that $\mathcal{U} \subseteq \mathcal{V}$. 
Assuming both families can model the label distribution, the task will at least as easy for the larger family.
This obviates the need to evaluate simpler function families (e.g., linear functions) when estimating dataset difficulty. 
Our experiments show that this bears out in practice as well (Appendix~\ref{appendix:roberta}).

\subsection{$\mathcal{V}$-Usable Information in Practice}
\label{sec:vinfo_experiments}

We consider the natural language inference (NLI) task, which involves predicting whether a text hypothesis entails, contradicts or is neutral to a text premise.
We first apply the $\mathcal{V}$-information framework to estimate the difficulty of a large-scale NLI dataset, Stanford NLI (SNLI; \citealp{bowman2015large}), across different state-of-the-art models.
The four models we use are GPT2-small \citep{radford2019language}, BERT-base-cased \citep{devlin2019bert}, DistilBERT-base-uncased \citep{sanh2019distilbert}, and BART-base \citep{lewis2020bart}.
Figure \ref{fig:vinfo-vs-accuracy} shows the $\mathcal{V}$-information estimate for all four, as well as their accuracy on the SNLI train and held-out (test) sets, across 10 training epochs. 
See Appendix~\ref{appendix:roberta} for results with larger models.

\paragraph{Model performance tracks $\mathcal{V}$-information.}
As seen in Figure \ref{fig:vinfo-vs-accuracy}, the model with the most $\mathcal{V}$-information on the SNLI test set is also the most accurate.
This is intuitive, since extracting more information makes prediction easier.
Overall, BART-base extracts the most $\mathcal{V}$-information, followed by BERT-base, DistilBERT-base, and GPT2-small; accuracy follows the same trend.

\paragraph{$\mathcal{V}$-information is more sensitive to over-fitting than held-out performance.}
At epoch 10, the $\mathcal{V}$-information is at its lowest for all models, although the SNLI test accuracy has only declined slightly from its peak, as seen in Figure \ref{fig:vinfo-vs-accuracy}.
This is because the models start becoming less certain about the correct label long before they start predicting the wrong label. 
This causes $H_\mathcal{V}(Y|X)$ to rise---and thus \vinfo to decline---even while most of the probability mass is still placed on the correct label.
This suggests that, compared to performance metrics like test accuracy, $\mathcal{V}$-information can more readily inform us of over-fitting.

\paragraph{Different datasets for the same task can have different amounts of $\mathcal{V}$-usable information.} 
We consider the MultiNLI dataset \cite{williams2018broad}, a multi-genre counterpart of SNLI.
Despite both being proxies for the NLI task, SNLI and MultiNLI have significantly different amounts of BERT-usable information, as shown in Figure \ref{fig:teaser}. 
The $\mathcal{V}$-information framework provides a principled means of measuring this difference in levels of difficulty; MultiNLI is expected to be more difficult than SNLI due to the diversity of genres it considers.
Also shown is CoLA \citep{warstadt2018neural}, a dataset for linguistic acceptability where each sentence is labeled as grammatical or not; this task is seemingly more difficult than NLI for BERT.

\section{Measuring Pointwise Difficulty}
\label{sec:pvi}

While $\mathcal{V}$-information provides an aggregate measure of dataset difficulty (\S\ref{sec:v-info}), a closer analysis requires measuring the degree of usable information in individual instances (w.r.t.\ a given distribution).
We extend the $\mathcal{V}$-information framework and introduce a new measure called \textbf{pointwise $\mathcal{V}$-information} (\pvi) for individual instances. 
The higher the \pvi, the easier the instance is for $\mathcal{V}$, under the given distribution.

\begin{definition}[Pointwise $\mathcal{V}$-Information]
Given random variables $X,Y$ and a predictive family $\mathcal{V}$, the pointwise $\mathcal{V}$-information (\pvi) of an instance $(x,y)$ is
\begin{equation}
   \text{\pvi}(x \to y) = -\log_2 g[\varnothing](y) + \log_2 g'[x](y)
\end{equation}
where functions $g = \arg\inf_{f \in \mathcal{V}} \mathbb{E} [- \log_2 f[\varnothing](Y) ]$ and $g' = \arg\inf_{f \in \mathcal{V}} \mathbb{E} [- \log_2 f[X](Y) ]$.
\end{definition}

If $\mathcal{V}$ were, for instance, the BERT function family, $g'$ and $g$ would be the models after finetuning BERT with and without the input respectively.
For a held-out instance $(x,y)$, $\text{\pvi}(x \to y)$ is the difference in the log-probability these models place on the gold label. 
\pvi is to $\mathcal{V}$-information what \pmi is to Shannon information: 
\begin{equation}
    \begin{split}
        I(X;Y) &= \mathbb{E}_{x,y \sim P(X,Y)}[\text{\pmi}(x,y)] \\
        I_\mathcal{V}(X \to Y) &= \mathbb{E}_{x,y \sim P(X,Y)}[\text{\pvi}(x \to y)]
    \end{split}
\end{equation}
Given this relationship, our understanding of $\mathcal{V}$-information extends to \pvi as well: higher \pvi instances are easier for $\mathcal{V}$ and vice-versa.
A higher \pvi increases the odds of being predicted correctly---this is intuitive because a correct prediction of a non-majority-class instance requires that some information be extracted from the instance.
Although the $\mathcal{V}$-information cannot be negative, the \pvi can be---much like how \pmi can be negative even though Shannon information cannot. 
A negative \pvi simply means that the model is better off predicting the majority class than considering $X$, which can happen for many reasons (e.g., mislabelling).
Examples with negative \pvi can still be predicted correctly, as long as $g'$ places most of the probability mass on the correct label.
Algorithm \ref{algo} shows our computation of \pvi and $\mathcal{V}$-information (by averaging over \pvi). 

The \pvi of an instance $(x,y)$ w.r.t.\ $\mathcal{V}$ should only depend on the distribution of the random variables. Sampling more from $P(X,Y)$ during finetuning should not change $\text{\pvi}(x \to y)$. 
However, an instance can be drawn from different distributions, in which case we would expect its \pvi to differ. 
For example, say we have restaurant reviews and movie reviews, along with their sentiment. 
The instance (``That was great!'', \emph{positive}) could be drawn from either distribution, but we would expect its \pvi to be different in each (even though $\mathcal{V}$ is the same).

\subsection{Implications}
\label{ssec:second_implications}

In addition to the comparisons that \vfamily-information allows us to make (\S\ref{sec:implications}), \pvi allows us to compare:
\begin{itemize}[noitemsep]
    \item[iv.] different instances $(x,y)$ by computing \pvi$(x \to y)$ for the same $X, Y, \mathcal{V}$ (Tables \ref{tab:hardest_and_easiest_cola}, \ref{tab:hardest_and_easiest}; Fig.~\ref{fig:instance_comparison})
    \item[v.] different slices or subsets of the data
    by computing the average \pvi over instances in each slice (Table \ref{tab:attributes_by_class}; Fig.~\ref{fig:slices}).
\end{itemize}
Note that the average \pvi of a slice of data is not its \vi, since we optimize the model w.r.t. the entire distribution.
However, since in practice one often wishes to understand the \textit{relative difficulty} of different subpopulations w.r.t.\ the training distribution, calculating the average \pvi---as opposed to the \vi of the subpopulation itself---is more useful.

\setcounter{algorithm}{-1}
\begin{algorithm}[t]
\footnotesize
\caption{\pvi and $\mathcal{V}$-Information}
\textbf{Input:} training data $\mathcal{D}_\text{train} = \{(\text{input}\ x_i, \text{gold label}\ y_i)\}_{i=1}^m$, held-out data $\mathcal{D}_\text{test} = \{(\text{input}\ x_i, \text{gold label}\ y_i)\}_{i=1}^n$, model $\mathcal{V}$ \\
\textbf{do}
\begin{algorithmic}

\STATE $g' \gets $ Finetune $\mathcal{V}$ on $\mathcal{D}_\text{train}$
\STATE $\varnothing \gets$ empty string (null input)
\STATE $g \gets $ Finetune $\mathcal{V}$ on $\{(\varnothing, y_i)\ |\ (x_i, y_i) \in \mathcal{D}_\text{train} \}$ \\
\STATE $H_\mathcal{V}(Y), H_\mathcal{V}(Y|X) \gets 0, 0$

\FOR{$(x_i, y_i) \in \mathcal{D}_\text{test}$}
\STATE $H_\mathcal{V}(Y) \gets H_\mathcal{V}(Y) - \frac{1}{n} \log_2 g[\varnothing](y_i)$
\STATE $H_\mathcal{V}(Y|X) \gets H_\mathcal{V}(Y|X) - \frac{1}{n} \log_2 g'[x_i](y_i)$
\STATE $\pvi(x_i \to y_i) \gets - \log_2 g[\varnothing](y_i) + \log_2 g'[x_i](y_i)$
\ENDFOR 

\STATE $\hat{I}_\mathcal{V}(X \to Y) = \frac{1}{n}\sum_i \pvi(x_i \to y_i) = H_\mathcal{V}(Y) - H_\mathcal{V}(Y|X)$
\end{algorithmic}
\textbf{end do}
\caption{After finetuning on a dataset of size $n$, the $\mathcal{V}$-information and \pvi can be calculated in $O(n)$ time.}
\label{algo}
\end{algorithm}

\subsection{\pvi in Practice}

\paragraph{\pvi can be used to find mislabelled instances.}
Correctly predicted instances have higher \pvi values than incorrectly predicted ones.
For the held-out sets in SNLI, MultiNLI and CoLA, the difference in mean \pvi between instances correctly and incorrectly predicted by BERT-base is 3.03, 2.87, and 2.45 bits respectively.
These differences are statistically significant ($p <$ 0.001).
Table \ref{tab:hardest_and_easiest_cola} shows the most difficult (lowest \pvi) instances from CoLA; we further find that some of these are in fact mislabelled (see Appendix~\ref{appendix:examples} for an analysis of SNLI). 
\begin{table}[t]
    \centering
    \footnotesize
    \resizebox{1\columnwidth}{!}{%
    \begin{tabular}{p{5.875cm}rr}
         \toprule                                                                  Sentence &  Label &    PVI \\
\midrule
                                                            \texttt{Wash you!} &      No & -4.616 \\
                             \texttt{Who achieved the best result was Angela.} &      No& -4.584 \\
                                            \texttt{Sue gave to Bill a book.} &      {No} & -3.649 \\
 \texttt{Only Churchill remembered Churchill giving the Blood, Sweat and Tears speech.} &      \textcolor{red}{No} & -3.571 \\
                                                    \texttt{Cynthia chewed.} &      \textcolor{red}{No} & -3.510 \\
                                                \texttt{It is a golden hair.} &      Yes & -3.251 \\
                                            \texttt{I won't have some money.} &      No & -3.097 \\
                \texttt{You may pick every flower, but leave a few for Mary.} &      \textcolor{red}{No} & -2.875 \\
\texttt{I know which book Mag read, and which book Bob said that you hadn't.} &      Yes & -2.782 \\
                                \texttt{John promise Mary to shave himself.} &      \textcolor{red}{Yes} & -2.609 \\
\bottomrule
    \end{tabular}
    }
    \caption{
    The 10 hardest (lowest \pvi) instances in the CoLA in-domain test set for grammaticality detection (label indicates grammaticality), according to BERT-base. 
    Examples in \textcolor{red}{red} are assessed to be mislabelled by authors of this work. 
    For e.g., `\texttt{Cynthia chewed.}' might be grammatical because the verb `chew' could be intransitive in this usage. 
    This suggests that \pvi could be used to identify mislabelled examples.
    All of these examples were predicted incorrectly by BERT-base.
    }
    \label{tab:hardest_and_easiest_cola}
\end{table}

\begin{figure}[t]
	\centering
	\includegraphics[width=\columnwidth]{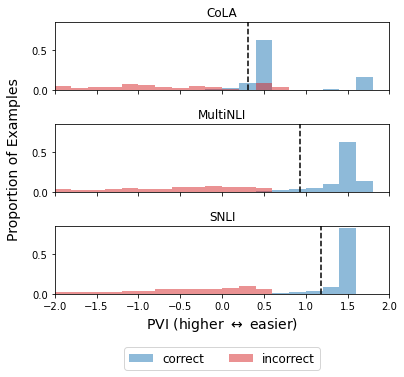}
	\caption{The distribution of \pvi for correctly and incorrectly predicted instances in each dataset.
	Note that the point at which instances start being incorrectly predicted is similar across datasets ($\sim$ 0.5 bits).
	In contrast, because the label space is different across CoLA and the other two datasets, such a comparison could not be made with a performance-based metric such as accuracy.
	}
	\label{fig:right_and_wrong}
\end{figure}

\paragraph{The \pvi threshold at which predictions become incorrect is similar across datasets.} 
In Figure \ref{fig:right_and_wrong}, we plot the \pvi distribution of correctly and incorrectly predicted instances in each dataset.
As expected, high-\pvi instances are predicted correctly and low-\pvi instances are not.
Notably, the point at which instances start being incorrectly predicted is similar across datasets (\pvi $\approx 0.5$).
Such a pattern could not be observed with a performance metric because the label spaces are different, evincing why the \vi framework is so useful for cross-dataset comparison.

\paragraph{\pvi estimates are highly consistent across models, training epochs, and random initializations.}
The cross-model Pearson correlation between \pvi estimates of SNLI instances is very high ($r >$ 0.80).
However, the cross-model Pearson correlation is lower for CoLA (0.40 $< r <$ 0.65); see Fig.~\ref{fig:correlations_heatmap} in Appendix~\ref{appendix:inter-epoch}.
This is because, as visualized in Figure \ref{fig:teaser}, CoLA has less usable information, making difficulty estimates noisier.
In the limit, if a dataset contained no usable information, then we would expect the correlation between \pvi estimates across different models and seeds to be close to zero.
It is also worth noting, however, that a high degree of cross-model correlation---as with SNLI---does not preclude comparisons between different models on the same dataset.
Rather, it suggests that in SNLI, a minority of instances are responsible for distinguishing one model's performance from another. 
This is not surprising---given the similar complexity and architecture of these models, we would expect most instances to be equally easy.
Moreover, despite the performance of Transformer-based models varying across random initializations \citep{dodge2019show,dodge2020fine,mosbach2020stability}, we find that \pvi estimates are quite stable: the correlation across seeds is $r > 0.85$ (for SNLI finetuned BERT-base, across 4 seeds); see Table \ref{tab:seed_corr} in Appendix~\ref{appendix:inter-epoch}.
It also concurs with human judgments of difficulty; see Fig.~\ref{fig:human_judgments} in Appendix~\ref{appendix:inter-epoch}.

\begin{figure}[t]
    \centering
    \includegraphics[width=\columnwidth]{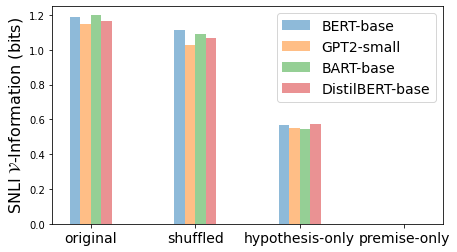}
    \caption{
    The amount of $\mathcal{V}$-usable information contained in different input attributes about the gold labels in SNLI. 
    The token identity alone (regardless of order) provides most of the information for all models (see \textsc{shuffled}).
    The \textsc{premise}, which can be shared by multiple instances, is useless alone; the \textsc{hypothesis}, which is unique to an instance, is quite useful even without a premise, suggesting it may contain annotation artefacts. 
    }
    \label{fig:attributes}
\end{figure}

\section{Uncovering Dataset Artefacts}
\label{sec:understanding}

A key limitation of standard evaluation metrics (e.g. accuracy) is the lack of interpretability---there is no straightforward way to understand \emph{why} a dataset is as difficult as it is.
$\mathcal{V}$-usable information offers an answer by allowing comparison of different input variables $X_i$ under the same $\mathcal{V}$ and $Y$, as implicated in \S\ref{sec:implications}.
We consider two approaches for this: applying input transformations (\S\ref{sec:transformation}), and slicing the dataset (\S\ref{sec:slicing}).

\subsection{Input Transformations}
\label{sec:transformation}

Our first approach involves applying different transformations $\tau_i(X)$ to isolate an attribute $a$, followed by calculating $I_\mathcal{V}(\tau_i(X) \to Y)$ to measure how much information (usable by $\mathcal{V}$) the attribute contains about the label. 
For example, by shuffling the tokens in $X$, we can isolate the influence of the word order attribute.

Given that a transformation may make information more accessible (e.g., decrypting some encrypted text; c.f.\ \S\ref{sec:v-info}), it is possible for $I_\mathcal{V}(\tau_i(X) \to Y) \geq I_\mathcal{V}(X \to Y)$, so the latter shouldn't be treated as an upper bound.
Such transformations were applied by \citet{o2021context} to understand what syntactic features Transformers use in next-token prediction; we take this a step further, aiming to discover annotation artefacts, compare individual instances, and ultimately understand the dataset itself.
We present our findings on SNLI, CoLA, as well as DWMW17 \citep{hateoffensive}, a dataset for hate speech detection, where input posts are labeled as hate speech, offensive, or neither.

We apply transformations to the SNLI input to isolate different attributes (see Appendix \ref{appendix:transformations} for an example): \emph{shuffled} (shuffle tokens randomly), \emph{hypothesis-only} (only include the hypothesis), \emph{premise-only} (only include the premise), \emph{overlap} (tokens in both the premise and hypothesis).
 
\paragraph{Token identity alone provides most of the usable information in SNLI.} 

Figure \ref{fig:attributes} shows that the token identity alone---isolated by shuffling the input---contains most of the usable information for all models. 
The premise, which is often shared by multiple instances, is useless alone; the hypothesis, which is unique to an instance, is useful even without a premise.
This corroborates the well-known annotation artefacts in SNLI \cite{gururangan2018annotation,poliak2018hypothesis}, which are spurious correlations exploited by models to predict the correct answer for the wrong reasons.

\paragraph{Hate speech detection might have lexical biases.}
Automatic hate speech detection is an increasingly important part of online moderation, but what causes a model to label speech as offensive?
We find that in DWMW17, the text contains 0.724 bits of BERT-usable information about the label.
Additionally, if one removed all the tokens, except for 50 (potentially) offensive ones---comprising common racial and homophobic slurs\footnote{These terms were manually chosen based on a cursory review of the dataset and are listed in Appendix \ref{appendix:transformations}.}---from the input post hoc, there still remains 0.490 bits of BERT-usable information.
In other words, just 50 (potentially) offensive words contain most of the BERT-usable information in DWMW17.
Our findings corroborate prior work which shows that certain lexical items 
(e.g., swear words, identity mentions) are responsible for hate speech prediction \cite{dixon18measuring,dinan2019build}.
Allowing models to do well by simply pattern-matching may permit subtleties in hate speech to go undetected, perpetuating harm towards minority groups \cite{blodgett2020language}.

\subsection{Slicing Datasets}
\label{sec:slicing}

\begin{table}[t]
    \centering
    \small
    \begin{tabularx}{\columnwidth}{Xrrr}
    \toprule
    {} &      Entailment &      Neutral &      Contradiction \\
    \midrule
 original        &  1.188 &  1.064 &  1.309 \\
shuffled    &  1.130 &  0.984 &  1.224 \\
hypothesis only  &  0.573 &  0.553 &  0.585 \\
premise only     &  0.032 & -0.016 & -0.016 \\
overlap &  0.415 &  0.177 &  0.298 \\
    \bottomrule
    \end{tabularx}
    \caption{The average amount of usable information (i.e., mean \pvi, in bits) that each attribute contains about each class in SNLI, according to BERT-base.
    Some attributes are more useful for a particular class: e.g., the degree of premise-hypothesis overlap is most useful for predicting entailment.
    Note that the mean \pvi for a particular class is different from the $\mathcal{V}$-information.
    }
    \label{tab:attributes_by_class}
\end{table}

\paragraph{Certain attributes are more useful for certain classes.} 
Comparing the usefulness of an attribute across classes can be useful for identifying systemic annotation artefacts.
This can be done by simply averaging the \pvi over the slice of data whose difficulty we are interested in measuring.
Note that the equivalence between \vi and expected \pvi only holds when the model used to estimate \pvi is trained over the entire dataset, which means that the average \pvi of a slice of data is not its \vi.
It would not make sense to estimate the \vi of a slice because it would require training on examples from just one class, in which case the \vi would be zero.
Thus the only usable difficulty measure is the mean \pvi.

We do this for SNLI in Table \ref{tab:attributes_by_class}.
We see that the tokens in the premise-hypothesis overlap contains much more BERT-usable information about the `entailment' class than `contradiction' or `neutral'.
This is unsurprising, given that the simplest means of entailing a premise is to copy it into the hypothesis and provide some additional detail.
If there is no inherent reason for an attribute to be more/less useful---such as overlap for entailment---there may be an artefact at work.
Even when there is an inherent reason for an attribute to be useful for a particular slice of the data, an attribute being exceptionally useful may also be evidence of a dataset artefact.
For example, if the premise-overlap hypothesis provided almost all the usable information needed for entailment, it may be because the crowdworkers who created the dataset took a shortcut by copying the premise to create the hypothesis. 

In Appendix \ref{appendix:instancewise}, we show how similar comparisons can be made between instances.

\begin{figure}[t]
    \centering
    \includegraphics[width=\columnwidth]{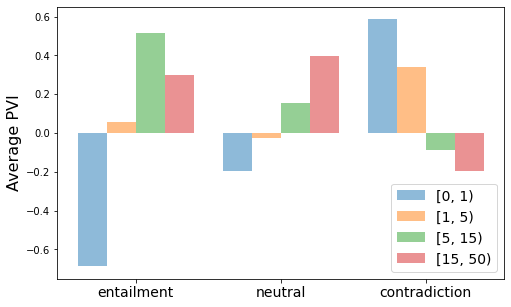}
    \caption{
    The mean \pvi of SNLI instances according to BERT-base, broken down by the overlap length (i.e., the number of tokens shared by the hypothesis and premise). Entailment examples with no overlap are the most difficult (i.e., lowest mean \pvi).
    }
    \label{fig:slices}
\end{figure}

\paragraph{Certain subsets of each class are more difficult than others.}
In Figure \ref{fig:slices}, we bin the examples in each SNLI class by the level of hypothesis-premise overlap and plot the average \pvi. 
We see entailment instances with no hypothesis-premise overlap are the most difficult (i.e., lowest mean \pvi) while contradiction instances with no overlap are the easiest (i.e., highest mean \pvi).
This is not surprising, since annotation artefacts in SNLI arise from constructing entailment and contradiction via trivial changes to the premise \cite{gururangan2018annotation}.

We additionally consider slices in the dataset based on dataset cartography \cite{swayamdipta2020dataset}, which uses training dynamics to differentiate instances via their (1) \emph{confidence} (i.e., mean probability of the correct label across epochs), and (2) \emph{variability} (i.e., variance of the former). 
The result is a dataset map revealing three regions: easy-to-learn, hard-to-learn, and ambiguous w.r.t.\ the trained model.
Slices of the dataset based on cartographic regions have distinct ranges of average \pvi (Fig.~\ref{fig:avg-pvi-carto}  in Appendix~\ref{sec:cartography}).

\subsection{Token-level Artefacts}

\begin{table}[t]
    \footnotesize

\textcolor{red}{WARNING: The following content contains language from the DWMW17 dataset that is offensive in nature.}

\centering
\begin{tabularx}{\columnwidth}{XXX}
    \toprule
    \multicolumn{3}{c}{DWMW17 \citep{hateoffensive}} \\
    \midrule
               Hate Speech &                Offensive &                 Neither \\
\midrule
 f*ggots (3.844) &   r*tards (2.821) &             lame (4.426) \\
      f*g (3.73) &      n*gs (2.716) &          clothes (0.646) \\
  f*ggot (3.658) &     n*gro (2.492) &              dog (0.616) \\
    c**ns (3.53) &       n*g (2.414) &              cat (0.538) \\
 n*ggers (3.274) &     c*nts (2.372) &  iDntWearCondoms (0.517) \\
\bottomrule
\end{tabularx}

\begin{tabularx}{\columnwidth}{XX}
    \toprule
    \multicolumn{2}{c}{CoLA \citep{warstadt2018neural}} \\
    \midrule
             Grammatical &             Ungrammatical \\
\midrule
will (0.267) &  book (2.737) \\
  John (0.168) &    is (2.659) \\
     . (0.006) &   was (2.312) \\
  and (-0.039) &    of (2.308) \\
    in (-0.05) &    to (1.972) \\
\bottomrule
\end{tabularx}

    \caption{Token-level annotation artefacts in DWMW17 and CoLA. These are the tokens whose omission leads to the greatest average increase in conditional entropy for each class (given in parentheses). Note that certain racial slurs are correctly identified as `hate speech' but in-group variants of the same terms fall under `offensive' instead. The full lists are available in Appendix \ref{appendix:artefacts}.}
    \label{tab:offensive_small}
\end{table}

Transforming $X$ and then measuring the $\mathcal{V}$-information to discover all token-level signals and artefacts is untenable, since we would need to finetune one new model per token.
Instead, we compute the change in the $\mathcal{V}$-information estimate after removing $t$, which yields modified input $x_{\neg t}$.
We use the same model $g'$ but evaluate only on a \textit{slice} of the data, $\mathcal{D}_{C,t}$, which contains the token $t$ and belongs to the class $C$ of interest.
This simplifies to measuring the increase in conditional entropy:
\begin{equation*}
\begin{split}
    \frac{1}{|\mathcal{D}_{C,t}|} \sum_{\mathcal{D}_{C,t}} [ & - \log_2 g'[x_{\neg t}](y) + \log_2 g'[x](y)]
\end{split}
\end{equation*}

\paragraph{Token-level signals and artefacts can be discovered using leave-one-out.} 
Table \ref{tab:offensive_small} shows that auxiliary verbs (e.g., be, did) and prepositions are most indicative of ungrammatical sentences in CoLA; in contrast, grammatical sentences have no strong indicators, with no word on average increasing the conditional entropy above 0.30 upon omission. 

In DWMW17, racial and homophobic slurs are the top indicators of `hate speech'.
However, in-group variants of the same racial slur---commonly used in African-American Vernacular English (AAVE)---fall under `offensive' instead.
The fact that AAVE terms are marked as `offensive' supports previous findings by that hate speech detection datasets may themselves be biased \citep{sap2019risk}.
In SNLI, we found many of the token-level artefacts matching those found using descriptive statistics in \citet{gururangan2018annotation}.
The complete word lists are available in Appendix \ref{appendix:artefacts}.

\subsection{Conditioning Out Information}

What if we wanted to measure how much BERT-usable information offensive words contain about the label in DWMW17 \textit{beyond} that which is captured in the sentiment?
In other words, if we already had access to the sentiment polarity of a text (positive/negative/neutral), how many \textit{additional bits of information} would the offensive words provide?
We cannot estimate this by simply subtracting $I_\mathcal{V}(\text{offensive} \to Y)$ from $I_\mathcal{V}(\text{sentiment} \to Y)$, since that difference could potentially be negative. Acquiring another random variable should not decrease the amount of information we have about the label (at worst, it should be useless).

To capture this intuition, \citet{hewitt2021conditional} proposed \textit{conditional \vi}, which allows one to condition out any number of random variables. Given a set of random variables $\mathcal{B}$ that we want to condition out, it is defined as:
\begin{equation}
    I_\mathcal{V}(X \to Y|\mathcal{B}) = H_\mathcal{V}(Y|\mathcal{B}) - H_\mathcal{V}(Y|\mathcal{B} \cup \{X\})
\end{equation}
The conditional entropy with respect to multiple variables is the only new concept here.
It is estimated in practice by concatenating the text inputs represented by $\mathcal{B}$ and $X$, which in our example is the sentiment polarity (one of `negative'/`neutral'/`positive') and the sequence of offensive words in the input.
The actual model family need not change to accommodate the longer text, as long as it remains under the input token limit.\footnote{If the inputs were vectors, the inputs represented by $\mathcal{B}$ and $\mathcal{B} \cup \{X\}$ would need to be of the same size; to do so, we would concatenate a zero vector to the former \citep{hewitt2021conditional}.}
We find that offensive words contain 0.482 bits of BERT-usable information about the label \emph{beyond} that which is contained in text sentiment.\footnote{The sentiment was categorized as positive/neutral/negative based on the polarity estimated by spaCy's built-in sentiment classifier \citep{spacy}.}
This is close to all of the BERT-usable information that the offensive words contain about the label (0.490 bits), suggesting that the predictive power of (potentially) offensive words is not mediated through sentiment in DWMW17.

\section{Related Work}
\label{sec:related}

While prior literature has acknowledged that not all data instances are equal \cite{vodrahalli2018all,swayamdipta2020dataset}, there have been few efforts to estimate dataset difficulty formally and directly.
As a notable exception, \citet{Zhang2020Dime} proposed DIME, an information-theoretic measure to estimate a lower bound on the lowest possible (i.e., model-agnostic) 0-1 error.
Model-agnostic approaches do not explain why some datasets are easier for some models, and have limited interpretability.
In contrast, $\mathcal{V}$-information and \pvi are specific to a model family $\mathcal{V}$.

Various techniques have been proposed to differentiate data instances within a dataset.
Text-based heuristics such as word identity \cite{bengio2009curriculum} or input length \cite{spitkovsky2010baby,gururangan2018annotation} have sometimes been used as proxies for instance difficulty, but offer limited insight into difficulty w.r.t.\ models.
Other approaches consider training loss \citep{han2018co,arazo2019unsupervised,shen2019learning}, confidence \citep{hovy-etal-2013-learning}, prediction variance \citep{chang2017active}, and area under the curve \citep{pleiss2020identifying}. 
Estimates relying on model training dynamics \citep{toneva2018empirical,swayamdipta2020dataset}, gradient magnitudes \cite{vodrahalli2018all}, or loss magnitudes \cite{han2018co} are sensitive to factors such as variance during steps of training.
Influence functions \citep{koh2017understanding}, forgetting events \citep{toneva2018empirical}, and the Data Shapley \citep{ghorbani2019data,jia2019towards} can all be used to assign pointwise estimates of importance to data instances based on their contribution to the decision boundary.
Moreover, although these methods all capture some aspect of difficulty, they do not lend themselves to interpreting datasets as readily as the predictive $\mathcal{V}$-information framework.

Given its dependence on training behavior across time, cartography \cite{swayamdipta2020dataset} offers complementary benefits to \vfamily-information.
It can be non-trivial to measure differences between, say a CoLA data map and an SNLI data map, w.r.t\ BERT.
In contrast, \vfamily-information provides a formal framework to make dataset difficulty estimates as an aggregate to compare datasets w.r.t\ a model.  
Other work has offered insight by splitting the data into ``easy'' and ``hard'' sets with respect to some attribute and studying changes in model performance, but these methods do not offer a pointwise estimate of difficulty \citep{sugawara2018makes,rondeau2018systematic,sen2020models}. 

Item response theory (IRT; \citealp{embretson2013item}) allows the difficulty of instances to be learned via  parameters in a probabilistic model meant to explain model performance \citep{lalor2018understanding,rodriguez-etal-2021-evaluation}. 
However, it does not formally relate dataset difficulty to the model being evaluated.
Estimating instance difficulty is also evocative of instance selection for active learning \citep{lewis1994heterogeneous,fu2013survey,liu2002issues}; however these estimates could change as the dataset picks up new instances.
In contrast, \pvi estimates are relatively stable, especially when the dataset  has higher $\mathcal{V}$-information.
Uncertainty sampling, for example, picks the instances that the partially trained model is least certain about \citep{lewis1994sequential,nigam2000text}, which could be interpreted as a measure of difficulty. 
However, once an instance is used for training, the model may become much more certain about it, meaning that the uncertainty values are unstable.

Interpretability of the role of certain attributes in trained models have lately led to the discovery of many dataset artefacts in NLP.
Our approach to discovering dataset artefacts can also complement existing approaches to artefact discovery \citep{gardner2021competency,pezeshkpour2021combining,le2020adversarial}.
Rissanen data analysis \citep{perez2021rissanen} offers a complimentary method for interpretability w.r.t\ attributes; it involves calculating the minimum description length (MDL): how many bits are needed to transmit the gold labels from a sender to a recipient when both have access to the same model and inputs.
Since the framework depends on the order of instances (i.e., what data has been transmitted thus far), it is unsuitable for estimating dataset difficulty.
In contrast, $\mathcal{V}$-information is defined w.r.t.\ a data distribution, so it is (in theory) agnostic to data and its ordering in fine-tuning.

$\mathcal{V}$-information \cite{xu2019theory} has had limited adoption in NLP.
It has been used to study what context features Transformers actually use \citep{o2021context}, as well as to condition out information for probing-based interpretability techniques \citep{hewitt2021conditional,pimentel2021bayesian}.
However, to the best of our knowledge, ours is the first approach to use $\mathcal{V}$-usable information for estimating the difficulty of NLP datasets.

\section{Future Work}
\label{sec:future}

There has been much work in the way of model interpretability, but relatively little in the way of dataset interpretability.
Our framework will allow datasets to be probed, helping us understand what exactly we test for in models and how pervasive annotation artefacts really are.
Moreover, our framework can be used proactively: by identifying the attributes responsible for dataset difficulty, one can create useful datasets out of otherwise useless raw data.
For example, the alignment of large language models is often bottlenecked by the lack of human preference data, which is slow and expensive to collect.
In Appendix \ref{sec:shp}, we show that starting with otherwise unremarkable web data, we can infer a dataset of collective preferences that contain as much usable information as preferences collected with paid human annotators, all the while being free and many times larger.

More immediate directions of future work include:
\begin{enumerate}
    \item Understanding how changes to the data distribution change the difficulty of individual examples.
   
    \item Extending \vi to open-ended text generation, which does not induce explicit distributions over the output space.
    This may requiring truncating the output space (e.g., using beam search with fixed width).
    
    \item Applying \vi to estimate dataset difficulty in other modalities (e.g., image, audio, tabular, etc.).
    There is nothing limiting the use of \vi to the NLP domain.
    For example, one could create a set of image filters---for different colors and objects---use them to transform the image, and then measure the drop in usable information.

\end{enumerate}

\section{Conclusion}
\label{sec:conclusion}

We provided an information-theoretic perspective to understanding and interpreting the difficulty of various NLP datasets.
We extended predictive $\mathcal{V}$-information to estimate difficulty at the dataset level, and then introduced pointwise $\mathcal{V}$-information (\pvi) for measuring the difficulty of individual instances.
We showed that instances with lower \pvi had lower levels of annotator agreement and were less likely to be predicted correctly.
We then demonstrated how systemic and token-level annotation artefacts in a dataset could be discovered by manipulating the input before calculating these measures.
Our studies indicate that $\mathcal{V}$-information offers a new, efficient means of interpreting NLP datasets.

\section*{Acknowledgements}

We thank Dan Jurafsky, Nelson Liu, Daniel Khashabi, and the anonymous reviewers for their helpful comments.
We thank Heidi (Chenyu) Zhang and Shabnam Behzad for helping create the SHP and SHP-2 datasets, alongside the first author, with advice from Dan Jurafsky and Yizhong Wang.
This project was supported by award DMS-2134012 from the NSF and the DARPA MCS program through NIWC Pacific (N66001-19-2-4031).


\bibliography{anthology,custom}
\bibliographystyle{icml2022}

\clearpage
\appendix
\onecolumn
\appendix

\section{Training Data}
\label{appendix:trainingdata}

\begin{figure}[h]
    \centering
    \includegraphics[width=0.5\columnwidth]{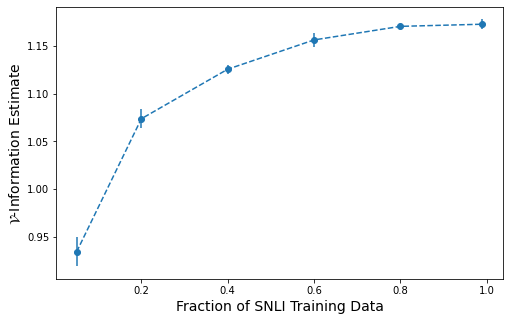}
    \caption{
    The $\mathcal{V}$-information estimate on the SNLI test set when BERT-base is trained on increasing fractions of the training data, drawn as a random sample (with replacement). 
    Here we plot the average and standard deviation across four samples for each fraction.
    }
    \label{fig:fractions}
\end{figure}

In Figure \ref{fig:fractions}, we plot the $\mathcal{V}$-information estimate on the SNLI test set as BERT-base is trained on increasing amounts of training data.
This is to test the assumption that the training set is sufficiently large to find the function $f \in \mathcal{V}$ that minimizes the conditional entropy.
Although this assumption is impossible to validate with complete certainty, since we do not have access to the true distribution, if the $\mathcal{V}$-information estimate plateaus before all the training data is used, it suggests that the training set size is not a limiting factor to the estimation.
We find that this is indeed the case with SNLI, where 80\% of the training data on averages provides the same estimate as using the entire training set.
In cases when this assumption does not hold, readers may want to consider measuring the Bayesian mutual information instead \citep{pimentel2021bayesian}.

\section{Larger Models}
\label{appendix:roberta}

\begin{figure*}[ht]
    \centering
    \includegraphics[width=\textwidth]{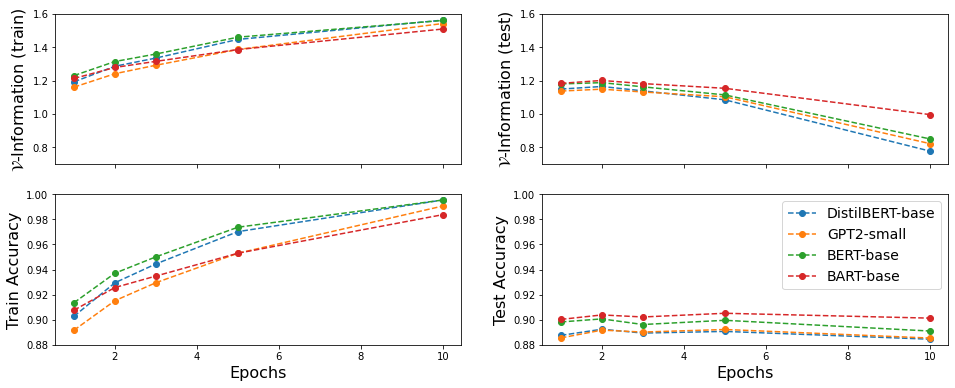}
    \caption{
    Comparing accuracy and the $\mathcal{V}$-information estimate on the SNLI train and test set w.r.t.\ various models.
    }
    \label{fig:snli_full}
\end{figure*}

\begin{figure*}[ht]
    \centering
    \includegraphics[width=\textwidth]{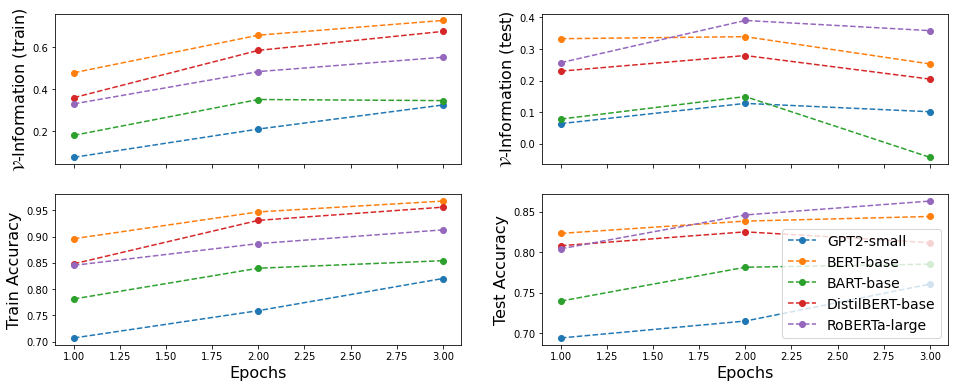}
    \caption{
    Comparing accuracy and the $\mathcal{V}$-information estimate on the CoLA in-domain train and held-out set w.r.t.\ various models.
    }
    \label{fig:cola}
\end{figure*}

In Figure \ref{fig:snli_full}, we plot the $\mathcal{V}$-information estimate for the SNLI test \emph{and} train sets.
In Figure \ref{fig:cola}, we plot the $\mathcal{V}$-information estimate on the CoLA in-domain held-out set for the four models that we previously studied, as well as a larger model, RoBERTa-large \citep{liu2019roberta}.
Despite the increase in scale, the trends observed in \S\ref{sec:v-info} still hold.

\section{Qualitative Analysis}
\label{appendix:examples}

\begin{table*}[h]
    \centering
    \footnotesize
    \begin{tabularx}{\textwidth}{p{6.5cm}p{6.5cm}cc}
        \toprule
       premise &                                                                                                       hypothesis &  label &    PVI \\
\midrule
                                                                                             Twenty five people are marching. &                                                                        A man plays the trombone on the sidewalk. &      N & -9.966 \\
                                                                                  A woman in a striped shirt holds an infant. &                                                                                         A person is watching TV. &      N & -9.612 \\
                                                                                        A person swimming in a swimming pool. &                                                                                       A person embraces the cold &      N & -9.152 \\
                                                                                       Women enjoying a game of table tennis. &                                                                                     Women are playing ping pong. &      E & -8.713 \\
                      A boy dressed for summer in a green shirt and kahki shorts extends food to a reindeer in a petting zoo. &                                                 A boy alien dressed for summer in a green shirt and kahki shorts &      \textcolor{red}{E} & -8.486 \\
                       Two skateboarders, one wearing a black t-shirt and the other wearing a white t-shirt, race each other. &                                                                                           Two snowboarders race. &      E & -8.087 \\
                                                                        An Asian woman dressed in a colorful outfit laughing. &                                                                                       The woman is not laughing. &      \textcolor{red}{E} & -7.903 \\
                                                          An older gentleman looks at the camera while he is building a deck. &  An older gentleman in overalls looks at the camera while he is building a stained red deck in front of a house. &      E & -7.709 \\
 A man wearing black pants, an orange and brown striped shirt, and a black bandanna in a "just thrown a bowling ball" stance. &                                                                                        The bandana is expensive. &      \textcolor{red}{C} & -7.685 \\
                                                     Two girls kissing a man with a black shirt and brown hair on the cheeks. &                                                                                                  Two girls kiss. &      C & -7.582 \\
\bottomrule

    \end{tabularx}
    \caption{The 10 hardest (lowest \pvi) instances in the SNLI test set, according to BERT-base. `E' denotes entailment, `N' neutral, and `C' contradiction. Instances that are possibly mislabelled are colored red.
    }
    \label{tab:hardest_and_easiest}
\end{table*}

In Table \ref{tab:hardest_and_easiest}, we list the 10 hardest instances in the SNLI test set according to BERT-base. All three classes---entailment, neutral, and contradiction---are represented in this list, with entailment being slightly over-represented. We see that some of the examples are in fact mislabelled---e.g., `PREMISE: An Asian woman dressed in a colorful outfit laughing. HYPOTHESIS: The women is not laughing.' is labelled as `entailment' even though the correct label is `contradiction'.

\section{Consistency of \pvi estimates}
\label{appendix:inter-epoch}
\begin{figure}[ht]
    \centering
    \includegraphics[width=0.8\textwidth]{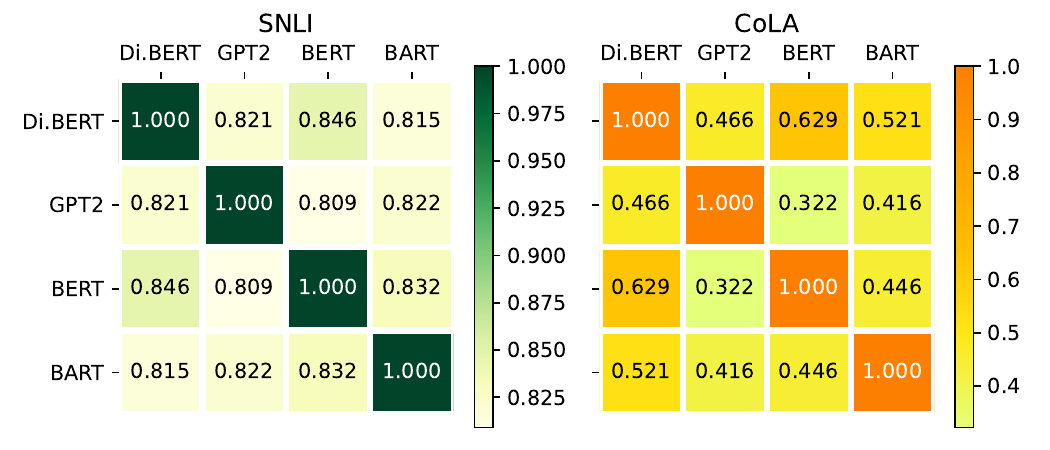}
    \caption{Cross-model Pearson's $r$ between \pvi estimates made by different finetuned models, on the SNLI and CoLA test sets. 
    For SNLI, the estimates are consistent: what one model finds difficult, others find difficult as well. 
    Since CoLA has less usable information for all these models, the correlations are lower.
    All correlations are highly statistically significant ($p < 0.001$).
   }
    \label{fig:correlations_heatmap}
\end{figure}

\begin{figure}[ht]
    \centering
    \includegraphics[width=0.5\textwidth]{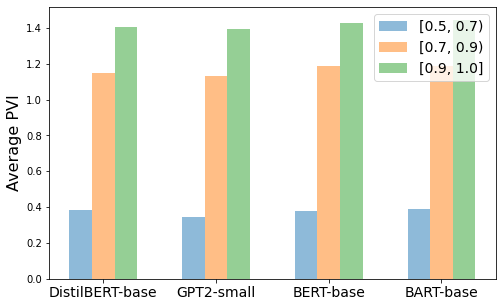}
    \caption{Examples that human annotators find easier (as measured by the fraction of annotators, in the range [0.5, 1.0], that agree with the gold label) also have higher \pvi on average.
   }
    \label{fig:human_judgments}
\end{figure}

\begin{table*}[t]
    \centering
    \small
    \begin{tabularx}{0.5\columnwidth}{Xrrrrr}
    \toprule
    \multicolumn{6}{c}{BERT-base} \\
    \midrule
    Epoch/Epoch &  1 &  2 &  3 &  5 &  10 \\
\midrule
1  &          1.000 &           0.908 &           0.871 &           0.838 &            0.762 \\
2  &          0.908 &           1.000 &           0.929 &           0.883 &            0.795 \\
3  &          0.871 &           0.929 &           1.000 &           0.879 &            0.796 \\
5  &          0.838 &           0.883 &           0.879 &           1.000 &            0.833 \\
10 &          0.762 &           0.795 &           0.796 &           0.833 &            1.000 \\
\bottomrule
    \end{tabularx}
    
    \begin{tabularx}{0.5\columnwidth}{Xrrrrr}
    \toprule
    \multicolumn{6}{c}{BART-base} \\
    \midrule
    Epoch/Epoch &  1 &  2 &  3 &  5 &  10 \\
\midrule
1   &          1.000 &           0.925 &           0.885 &           0.853 &            0.754 \\
2  &          0.925 &           1.000 &           0.952 &           0.906 &            0.807 \\
3  &          0.885 &           0.952 &           1.000 &           0.914 &            0.814 \\
5  &          0.853 &           0.906 &           0.914 &           1.000 &            0.862 \\
10 &          0.754 &           0.807 &           0.814 &           0.862 &            1.000 \\
\bottomrule
    \end{tabularx}

    \begin{tabularx}{0.5\columnwidth}{Xrrrrr}
    \toprule
    \multicolumn{6}{c}{DistilBERT-base} \\
    \midrule
    Epoch/Epoch &  1 &  2 &  3 &  5 &  10 \\
\midrule
1   &          1.000 &           0.928 &           0.884 &           0.828 &            0.766 \\
2  &          0.928 &           1.000 &           0.952 &           0.890 &            0.825 \\
3  &          0.884 &           0.952 &           1.000 &           0.900 &            0.819 \\
5  &          0.828 &           0.890 &           0.900 &           1.000 &            0.860 \\
10 &          0.766 &           0.825 &           0.819 &           0.860 &            1.000 \\
\bottomrule
    \end{tabularx}
    
        \begin{tabularx}{0.5\columnwidth}{Xrrrrr}
    \toprule
    \multicolumn{6}{c}{GPT2} \\
    \midrule
    Epoch/Epoch &  1 &  2 &  3 &  5 &  10 \\
\midrule
1   &          1.000 &           0.931 &           0.887 &           0.855 &            0.747 \\
2  &          0.931 &           1.000 &           0.961 &           0.918 &            0.813 \\
3  &          0.887 &           0.961 &           1.000 &           0.933 &            0.827 \\
5  &          0.855 &           0.918 &           0.933 &           1.000 &            0.874 \\
10 &          0.747 &           0.813 &           0.827 &           0.874 &            1.000 \\
\bottomrule
    \end{tabularx}

    \caption{
    Cross-epoch Pearson correlation between \pvi estimates made on the SNLI test set while finetuning various models on the SNLI training set. 
    The estimates are stable: when an instance is easy(difficult) early on, it generally remains easy(difficult).
    For all models studied, the cross-epoch correlation does not dip below 0.80 for the first five epochs.
    }
    \label{tab:intra_corr}
\end{table*}

\begin{table}[ht]
    \centering
    \small
    \begin{tabularx}{0.5\columnwidth}{Xrrrr}
        \toprule
        seed &   Run 1 &  Run 2 &  Run 3 &  Run 4 \\
        \midrule
        Run 1 &                 1.000 &                 0.877 &                 0.884 &                 0.885 \\
        Run 2 &                 0.877 &                 1.000 &                 0.887 &                 0.882 \\
        Run 3 &                 0.884 &                 0.887 &                 1.000 &                 0.895 \\
        Run 4 &                 0.885 &                 0.882 &                 0.895 &                 1.000 \\
        \bottomrule
        \end{tabularx}
    \caption{Cross-model Pearson correlation between \pvi estimates made after one epoch of finetuning BERT on SNLI with different seeds. 
    Estimates are stable: what a model finds difficult is mostly not due to chance.
    }
    \label{tab:seed_corr}
\end{table}

\paragraph{Cross-Model Correlations} Figure~\ref{fig:correlations_heatmap} shows a heatmap for Cross-model Pearson's $r$ between \pvi estimates made by different finetuned models, on the SNLI and CoLA test sets; these results support the findings in \S\ref{sec:vinfo_experiments}.

\paragraph{Human Agreement} In Figure \ref{fig:human_judgments}, we plot the average \pvi at different levels of annotator agreement. 
We find that there is a concurrence between what humans find difficult and what examples are difficult according to \pvi.

\paragraph{Cross-Epoch Correlations} In Table \ref{tab:intra_corr}, we list the cross-epoch Pearson correlation between \pvi estimates made by the same model on the SNLI test set over the course of finetuning. The correlation is high ($r > 0.80$ during the first 5 epochs), suggesting that when an instance is easy(difficult) early on, it tends to remain easy(difficult).

\paragraph{Cross-Seed Correlations} In Table \ref{tab:seed_corr}, we list the Pearson correlation between \pvi estimates made by BERT across different training runs. The correlation is high ($r > 0.87$), suggesting that what a model finds difficult is not due to chance.

\section{Transformations}
\label{appendix:transformations}

\begin{table*}[ht]
    \small
    \centering
    \begin{tabularx}{\textwidth}{llX}
    \toprule
        Attribute & Transformation & Transformed Input \\
    \midrule
        Original & & PREMISE: Two girls kissing a man with a black shirt and brown hair on the cheeks. HYPOTHESIS: Two girls kiss. \\
        Shuffled & shuffle tokens randomly & PREMISE: girls two a kissing man with a black cheeks shirt and hair brown on the . HYPOTHESIS: kiss two . girls \\
        Hypothesis-only & only include hypothesis & HYPOTHESIS: Two girls kiss. \\
        Premise-only & only include premise & PREMISE: Two girls kissing a man with a black shirt and brown hair on the cheeks.\\
        Overlap & hypothesis-premise overlap & PREMISE: Two girls [MASK] [MASK] [MASK] [MASK] [MASK] [MASK] [MASK] [MASK] [MASK] [MASK] [MASK] [MASK] [MASK] . HYPOTHESIS: Two girls [MASK] . \\
    \bottomrule
    \end{tabularx}
    \caption{Given an NLI instance (see `Original'), each transformation isolates some attribute from the input. 
    The headers `PREMISE' and `HYPOTHESIS' were added by us to transform the two sentence inputs into a single text input for all models that were evaluated.
    \longtermtodos{Not sure these are the best transformations for the length and overlap attributes; try something simple by concatenating the prem-hyp pair with an bucketed value of the attribute (similar to overlap). \kawin{but that doesn't isolate the effect of an attribute right? if the model is complex enough to extract the overlap from the original data, then the original and transformed v-info will be the same}} 
    }
    \label{tab:transformations}
\end{table*}

\textcolor{red}{WARNING: The following content contains language from the DWMW17 dataset that is offensive in nature.}

In Table \ref{tab:transformations}, we provide an instance from the SNLI test set in its original form and after various attribute-specific transformations have been applied to it.
These only capture a small subset of the space of possible transformations.

For DWMW17, we hand-picked a set of 50 potentially offensive words based on a cursory review of the dataset to see how much information these terms alone contain about the label: `nigga', `niggas', `niggah', `niggahs', `hoe', `hoes', `bitch', `bitches', `whitey', `white trash', `cracker', `crackers', `beaner', `beaners', `pussy', `pussies', `fag', `fags', `faggot', `faggots', `ho', `hos', `redneck', `rednecks', `porn', `fuck', `fucks', `fucker', `fuckers', `motherfucker', `motherfuckers', `nigger', `niggers', `coon', `coons', `niggaz', `nig', `nigs', `slut', `sluts', `wigger', `wiggers', `fucked', `fucking', `wigga', `wiggas', `retard', `retards', and `retarded'.

\longtermtodos{in addition to using these terms, perhaps consider using already curated lists. See \citet{zhou2021challenges}.}

\section{Instance-wise Comparisons}
\label{appendix:instancewise}

\begin{figure}[t]
    \footnotesize
    \centering
    \textcolor{black}{\#7717: \emph{PREMISE: Little kids play a game of running around a pole. HYPOTHESIS: The kids are fighting outside.}} \\
    \textcolor{black}{\#9627: \emph{PREMISE: A group of people watching a boy getting interviewed by a man. HYPOTHESIS: A group of people are sleeping on Pluto.}}
    
    \includegraphics[width=0.5\columnwidth]{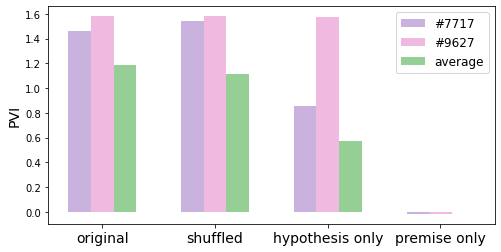}
    \caption{
    The \pvi of two SNLI `neutral' instances (\#7717 and \#9627) w.r.t.\ BERT-base after attribute-specific transformations, as well as the $\mathcal{V}$-information estimate (i.e., average \pvi over the data) for each attribute. 
    The latter instance is easier for BERT, likely because its hypothesis is much more informative due to being so different from its premise.
    Note that it makes sense to compare instances w.r.t.\ the same attribute, but not different attributes w.r.t.\ the same instance, since the models used to estimate the attribute $\mathcal{V}$-information $I_\mathcal{V}(\tau_a(X) \to Y)$ are chosen to maximize the likelihood of \textit{all the data}.
    }
    \label{fig:instance_comparison}
\end{figure}

Certain attributes are responsible for the difficulty of certain examples. Figure \ref{fig:instance_comparison} is an example of how we might do a fine-grained comparison of instances to understand why one may be more difficult for a given model.
We compare two SNLI `neutral' instances from the test set to try to understand why \#9627 is easier for BERT than \#7717 (i.e., why $\pvi(x_{9627} \to y_{9627}) > \pvi(x_{7717} \to y_{7717})$), finding that it is likely due to the former's \emph{hypothesis} being more informative.
While different instances can be compared w.r.t.\ the same attribute, different attributes cannot be compared w.r.t.\ the same instance, since the models used to estimate the attribute-specific $\mathcal{V}$-information $I_\mathcal{V}(\tau_a(X) \to Y)$ are chosen to maximize the likelihood of \emph{all the data}.
This is why, for example, the \pvi of \#7717 is higher after its tokens have been shuffled even though the average \pvi (i.e., dataset-level $\mathcal{V}$-information) declines after shuffling tokens.

\section{Token-Level Artefacts}
\label{appendix:artefacts}

\textcolor{red}{WARNING: The following content contains language from the DWMW17 dataset that is offensive in nature.}
In Table \ref{tab:token_artefacts}, we list the tokens in the SNLI, CoLA, and DWMW17 datasets that, when dropped out, cause the greatest decrease in the $\mathcal{V}$-information estimate.
These are token-level artefacts of each class in the dataset.
In the DWMW17 hate speech detection dataset, racial and homophobic slurs are artefacts of hate speech, while ableist and sexual slurs are artefacts of offensive speech.
In-group AAVE terms are also predictive of offensive speech in DWMW17 even when they are used non-offensively, hinting at possible bias in the dataset \citep{sap2019risk}.
In CoLA, auxiliary verbs and prepositions are artefacts of ungrammatical sentences; grammatical sentences don't have any artefacts.
For SNLI, we recover many of the token-level artefacts found by \citet{gururangan2018annotation} using descriptive statistics---even uncommon ones, such as `cat' for contradiction.

\section{Relation to Dataset Cartography}
\label{sec:cartography}

\begin{figure}[t]
    \centering
    \begin{minipage}{0.49\textwidth}
        \centering
        \includegraphics[width=0.95\textwidth]{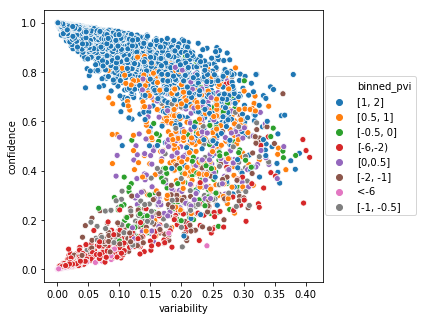}
        \caption{Relationship between \pvi and the training dynamics-based data map \citep{swayamdipta2020dataset} for SNLI held-out (test) set, computed for the DistilBERT-base architecture.
    As in \citet{swayamdipta2020dataset}, $Y$-axis corresponds to \textit{confidence}, i.e. the mean probabilities of the true class across training epochs, and $X$-axis corresponds to \textit{variability}, i.e. the standard deviation of the true class probabilities across the same.
    Colours indicate binned values of \pvi.
    \pvi estimates track closely with \textit{confidence}.}
    \label{fig:cartography}
    \end{minipage}\hfill
    \begin{minipage}{0.49\textwidth}
        \centering
        \includegraphics[width=0.95\textwidth]{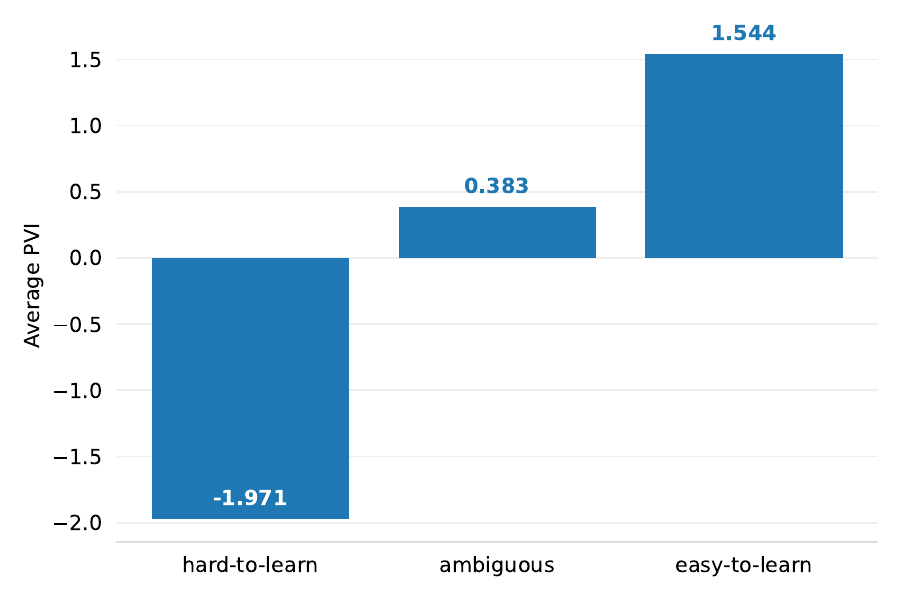} 
        \caption{Average \pvi of the 10\% of the most ambiguous, and the 10\% of the hardest-to-learn, and 10\% of the easiest-to-learn regions of the SNLI / DistilBERT-base data map (Fig. \ref{fig:cartography}).
        Hard-to-learn instances are frequently mislabeled, and therefore also reflect the lowest average \pvi values. 
        The highest average \pvi values are possessed by the easy-to-learn instances, which are the most common class of instances in the SNLI dataset.
        Ambiguous instances are those that the model changes it decision on frequently through training; these correspond to lower average \pvi values than easy-to-learn instances.}
        \label{fig:avg-pvi-carto}
    \end{minipage}
\end{figure}

\citet{swayamdipta2020dataset} introduced dataset cartography, a method to automatically analyze and diagnose datasets with respect to a trained model.
It offers a complimentary understanding of datasets and their properties, taking into account the behavior of a model towards different data instances during training. 
This behavior---training dynamics---helps differentiate instances via their (1) \emph{confidence} (i.e., mean probability of the correct label across epochs), and (2) \emph{variability} (i.e., variance of the former). 
The result is a dataset map revealing three regions:
\begin{itemize}
    \item \textbf{Easy-to-learn} (high \textit{confidence}, low \textit{variability}) instances are the most frequent instances in the dataset, those which high capacity models like BERT predict correctly throughout training.
    \item \textbf{Hard-to-learn} (low \textit{confidence}, low \textit{variability}) instances correspond to those which are predicted incorrectly throughout training; these were shown to correspond to mislabeled examples often.
    \item \textbf{Ambiguous} (high \textit{variability}) instances correspond to those which the model often changes its prediction for; these examples are the ones which are most responsible for high test performance, both in and out of distribution.
\end{itemize}

Figure ~\ref{fig:cartography} shows that \pvi values track closely to the confidence axis of a SNLI-DistilBERT-base data map\footnote{Data maps were originally plotted on training data; however, they can be plotted on held-out data by computing training dynamics measures on the same, after every training epoch.} \cite{swayamdipta2020dataset}.
Data maps and \pvi estimates offer orthogonal perspectives to instance difficulty, the former capturing behavior of instances as training proceeds.
Moreover, \vfamily-information can estimate dataset difficulty as an aggregate (\S\ref{sec:v-info}), which is not the case for training dynamics metrics, which offer only point estimates.
Both approaches can be helpful for discovering data artefacts.
Predictive \vfamily-information estimates, however, offer the unique capability of transforming the input to discover the value of certain attributes in an efficient manner.

In Figure \ref{fig:avg-pvi-carto}, we report the average \pvi estimates of the three regions discovered via data maps:
\begin{itemize}
    \item \textbf{Easy-to-learn} (high \textit{confidence}, low \textit{variability}) instances correspond to the highest average \pvi, indicating that they have the highest amount of DistilBERT-usable information.
    \item \textbf{Hard-to-learn} (low \textit{confidence}, low \textit{variability}) instances correspond to the lowest average \pvi, indicating that they have the lowest amount of DistilBERT-usable information. This is not surprising, since they also correspond to mislabeled instances, which can be difficult to extract usable information from.
    \item \textbf{Ambiguous} (high \textit{variability}) instances correspond to lower average \pvi, indicating that there is some usable information, but not as much as those of the easy-to-learn instances, w.r.t\ DistilBERT.
\end{itemize}
For each of the bars in the plot, we consider 10\% of the dataset belonging to each region (with the highest corresponding measures of \textit{confidence} and \textit{variability}).
\section{Creating Datasets}
\label{sec:shp}

$\mathcal{V}$-information can be used to not only understand datasets but to create them anew.
For example, consider the problem of aligning large language models (LLMs) with human preferences \citep{ouyang2022training}, so as to make them more helpful to users while limiting potential harm.
This task is often bottlenecked by the lack of human preference data, which is slow and expensive to collect, as it relies on human annotation.
In theory, online fora such as Stack Exchange and Reddit offer a free source of preferences: users collectively vote on comments in response to a question and the aggregate scores---the number of up-votes net of down-votes---ostensibly reflect a group preference.
This gives us tuples $(q, r_A, r_B, y)$, where $q$ is a question (e.g., \textit{How do I salt food?}), $r_A$ and $r_B$ are two root-level answers, and $y$ is 1 if $r_A$ has a higher aggregate score than $r_B$ and 0 otherwise.

However, in their raw form, such data contain no usable information: $I_\mathcal{V}(\{Q, R_A, R_B\} \to Y) \approx 0$, even for capacious model families such as \texttt{GPT-3}.
But why?
We find that the examples with the highest \pvi (i.e., the most learnable) are those where the higher-scoring comment was written \textit{after} the lower-scoring one.
In retrospect, this is intuitive: the earlier a comment is written, the more time it has to accrue votes, and since most comments end up with a positive score---up-voting being much more common than down-voting---an earlier comment is much more likely to have a higher score than a later one.
If $r_A$ has a higher score than $r_B$ but was written earlier, we do not whether it is genuinely more preferred or simply benefits from greater visibility.
However, if $r_A$ has a higher score despite being written later, it suggests that it was so preferred by users that it was able to overcome a visibility disadvantage, meaning that it is more likely to reflect a genuine preference.
By correcting for this confound and keeping only the tuples where the higher-scoring comment was written after the lower-scoring one, we get a dataset with as much \texttt{GPT-3}-usable information ($\approx$ 0.50 bits) as HH-RLHF \citep{ganguli2022red}, one of the most widely used preference datasets in LLM alignment.

We use this filter, as well as a handful of others, to create two of the largest datasets of collective human preferences over text, which we call the Stanford Human Preferences datasets (SHP and SHP-2).
SHP contains 385K preferences inferred from Reddit data, specifically advice-oriented subreddits in a set of hand-curated subject areas (e.g., \texttt{askculinary}). 
SHP-2 contains 4.8M preferences from both Reddit and various Stack Exchange domains (e.g., \texttt{stackoverflow}).
For licensing information, schema specification, potential biases, and more details on how the specific subreddits and domains were selected, we refer the reader to our Huggingface repositories for \href{https://huggingface.co/datasets/stanfordnlp/SHP}{SHP} and \href{https://huggingface.co/datasets/stanfordnlp/SHP}{SHP-2}.

\begin{table*}[t]
    \centering
    \footnotesize
    \textcolor{red}{WARNING: The following content contains language from the DWMW17 dataset that is offensive in nature.}

\begin{tabularx}{\textwidth}{XXX}
    \toprule
    \multicolumn{3}{c}{DWMW17 \citep{hateoffensive}} \\
    \midrule
               Hate Speech &                Offensive &                 Neither \\
\midrule
 f*ggots (3.844) &   r*tards (2.821) &             lame (4.426) \\
      f*g (3.73) &      n*gs (2.716) &          clothes (0.646) \\
  f*ggot (3.658) &     n*gro (2.492) &              dog (0.616) \\
    c*ons (3.53) &       n*g (2.414) &              cat (0.538) \\
 n*ggers (3.274) &     c*nts (2.372) &  iDntWearCondoms (0.517) \\
   qu*er (3.163) &    p*ssies (2.29) &             thank (0.47) \\
    co*n (3.137) &     qu*er (2.213) &             kick (0.423) \\
  n*gger (3.094) &  r*tarded (1.997) &               30 (0.345) \\
     d*ke (3.01) &      c*nt (1.919) &         football (0.334) \\
    f*gs (2.959) &   b*tches (1.858) &             soul (0.323) \\
\bottomrule
\end{tabularx}

\begin{tabularx}{\textwidth}{XXX}
    \toprule
    \multicolumn{3}{c}{SNLI \citep{bowman2015large}} \\
    \midrule
              Entailment &                 Neutral &                  Contradiction \\
\midrule
         nap (3.256) &        tall (4.246) &      Nobody (7.258) \\
        bald (3.183) &       naked (2.193) &         not (4.898) \\
      crying (2.733) &     indoors (1.724) &          no (4.458) \\
       Woman (2.517) &       light (1.442) &       naked (3.583) \\
      asleep (2.482) &         fun (1.318) &      crying (2.938) \\
    sleeping (2.416) &         bed (1.006) &     indoors (2.523) \\
        soda (2.267) &  motorcycle (0.993) &  vegetables (2.295) \\
         bed (2.136) &       works (0.969) &    sleeping (2.293) \\
         not (2.111) &        race (0.943) &      jogging (2.17) \\
 snowboarder (2.099) &    daughter (0.924) &         cat (2.092) \\

\bottomrule
\end{tabularx}

\begin{tabularx}{\textwidth}{XX}
    \toprule
    \multicolumn{2}{c}{CoLA \citep{warstadt2018neural}} \\
    \midrule
             Grammatical &             Ungrammatical \\
\midrule
will (0.267) &  book (2.737) \\
  John (0.168) &    is (2.659) \\
     . (0.006) &   was (2.312) \\
  and (-0.039) &    of (2.308) \\
    in (-0.05) &    to (1.972) \\
    ' (-0.063) &   you (1.903) \\
   to (-0.195) &    be (1.895) \\
   of (-0.195) &    in (1.618) \\
 that (-0.379) &   did (1.558) \\
  the (-0.481) &   The (1.427) \\
\bottomrule
\end{tabularx}

    \caption{Token-level annotation artefacts in each dataset. These are the tokens whose omission leads to the greatest average increase in conditional entropy for each class (given in parentheses).}
    \label{tab:token_artefacts}
\end{table*}


\end{document}